\pdfoutput=1

\documentclass[11pt, table,xcdraw]{article}
\usepackage[]{EMNLP2023}

\usepackage{times}
\usepackage{latexsym}
\usepackage{multirow}
\usepackage{hyperref}
\usepackage{float}
\usepackage[T1]{fontenc}

\usepackage[utf8]{inputenc}

\usepackage{microtype}

\usepackage{inconsolata}
\usepackage{graphics}
\usepackage{graphicx}

\title{Argumentative Stance Prediction: An Exploratory Study on Multimodality and Few-Shot Learning}

\author{Arushi Sharma\thanks{Equal contribution} , Abhibha Gupta\footnotemark[1] , Maneesh Bilalpur\footnotemark[1] \\
School of Computing and Information \\
University of Pittsburgh\\
\texttt{\{arushi.sharma, abg96, mab623\}@pitt.edu}
}

\begin{document}
\maketitle
\begin{abstract}
%



To advance argumentative stance prediction as a multimodal problem, the \textit{First Shared Task in Multimodal Argument Mining} hosted stance prediction in crucial social topics of gun control and abortion. Our exploratory study attempts to evaluate the necessity of images for stance prediction in tweets and compare out-of-the-box text-based large-language models (LLM) in few-shot settings against fine-tuned unimodal and multimodal models. Our work suggests an ensemble of fine-tuned text-based language models (0.817 F1-score) outperforms  both the multimodal (0.677 F1-score) and text-based few-shot prediction using a recent state-of-the-art LLM (0.550 F1-score). In addition to the differences in performance, our findings suggest that the multimodal models tend to perform better when image content is summarized as natural language over their native pixel structure and, using in-context examples improves few-shot performance of LLMs.


\end{abstract}

\section{Introduction}

Argumentative stance studies related to ideological topics offer valuable insights into complex dynamics of opinion, belief and discourse in various domains. These insights have far-reaching implications, extending their influence over areas including public opinion, social dynamics, and policy efficacy. By predicting the stance in real-time, policymakers and stakeholders can get immediate feedback on public reaction to new proposals or laws, allowing them to make timely and informed decisions.

Argumentative stance prediction is becoming a major endeavor in multiple research fields as the reliance on sentiment detection may be sub-optimal  \cite{reveilhac2023replicable}. While the stance prediction task appears similar to sentiment analysis, it has many theoretical differences. Sentimental analysis primarily focuses on emotions, whereas the stance prediction need not necessarily coincide with the sentiment directed towards the target. Stance prediction for sensitive and polarizing topics can be more challenging, particularly within the brief context of informal social media text \cite{alturayeif2023systematic}.

Previous studies have primarily concentrated on examining stance prediction in textual modalities \cite{alturayeif2023systematic, hosseinia2020stance}. However, an increasing number of recent works are widening the focus to include other modalities, such as images. Since multimodality helps us understand language from the modalities of text, vision and acoustic, \cite{zadeh2018proceedings}, the application of multimodal inputs in argumentative stance prediction seems promising.

Towards the perpetuation of multimodality in argumentative stance prediction as a part of the ImgArg 2023 \cite{liu-etal-2023-overview} challenge, we explore the following questions using a dataset of tweets on gun control and abortion topics:

\begin{enumerate}
    \item How well does language as a stand-alone modality perform at argumentative stance prediction?
    \item Does incorporating image information improve prediction performance?
    \item How do Large-Language Models (LLMs) in few-shot setting compare against fine-tuned unimodal and multimodal models?
\end{enumerate}

Our work shows that an ensemble of fine-tuned language models performs the best for argumentative stance prediction from tweets. Incorporating image information into text using state-of-the-art multimodal models does not outperform the ensemble model. LLMs (particularly, LLaMA-2) in few-shot setting exhibit high recall but suffer from low precision. Though using in-context examples in few-shot setting improves performance, they underperform the ensemble model. 

\section{Related Work}


Existing work has explored the interplay between stance and sentiment to enhance stance detection. \cite{sobhani2017stance} investigated the relationship between stance and sentiment, utilizing SVM with N-gram, word embedding, and sentiment lexicon features. They concluded that while sentiment features offer utility, they are insufficient on their own for effective stance detection. Meanwhile, \cite{hosseinia2020stance} showcased the prowess of bidirectional transformers in achieving competitive performance without fine-tuning, harnessing sentiment and emotion lexicons. Their findings show the efficacy of sentiment information, as opposed to emotion, in discerning the stance.


\cite{alturayeif2023systematic} conducted an extensive analysis of 96 primary studies spanning eight machine learning techniques for stance detection and its applications. The analysis suggests that deep learning models with self-attention mechanisms were found to be frequently outperforming the traditional machine learning models such as SVM, and emerging techniques like few-shot learning and multitask learning were increasingly applied for stance detection.

Multimodal stance detection is being increasingly used for  social applications such as rumor verification \cite{Zhang2021MultiModalMM} and identifying public attitudes towards climate change on Twitter \cite{Upadhyaya2023AMM}. Despite recent advancements in multimodal language models \cite{Wang2023LargescaleMP}, the use of image modality for stance detection remains an underexplored area. Our work conducts an exploratory study to investigate the necessity of multimodal models for stance detection and compares different ways to incorporate image information into text modality.

\section{Dataset and Task}
The ImgArg dataset \cite{liu2022imagearg} is a part of the \textit{Multimodal Argument Mining} \cite{liu-etal-2023-overview} competition. Curated with the goal of expanding argumentation mining into multimodal realm, the dataset consists of Twitter texts along with their images from two topics--gun control and abortion. Each text-image pair corresponding to a tweet are annotated with a stance (support or oppose) along with its persuasiveness (no persuasiveness to extremely persuasive). In this paper, we focus on the stance prediction task. Briefly, the task can be described as given an image-text pair corresponding to a tweet, predict if it supports or opposes the topic.


It is important to note that while the gun control dataset is balanced, the abortion dataset is imbalanced by a 1:3 support:oppose stance ratio. The gun control and abortion training sets are 920\footnote{The organizers reported 923 tweets, however, three tweets were dropped because of download issues.} and 891 tweets respectively. Both datasets have an equal number of tweets in the validation (100 tweets) and test (150 tweets) sets. 


\section{Approach}
To predict argumentative stance over multimodal tweets from gun control and abortion topics, we leverage three different ideas. We explore an ensemble of LLMs against its constituent models, incorporate image information through multimodal models as well as evaluate out-of-the-box LLMs in few-shot setting. This section describes the experimental approaches used in the process. Further details can be found in the appendix.


\subsection{Ensemble Stance Prediction}
Individual language models have demonstrated their superior performance across a variety of tasks. However, ensemble methods tend to perform better \cite{jiang2023llm} than their constituent models. To explore this idea, we evaluated text-based language models such as XLNet \cite{yang2019xlnet}, XLM-RoBERTa \cite{conneau2019xlm-roberta}, Transformer XL \cite{dai2019transformer-xl}, DeBERTa-v2 \cite{he2020deberta}, BLOOM-560M \cite{scao2022bloom}. 
Since the dataset is a collection of tweets, conventional problems such as very long sequence length were non-existent.

Ensemble decisions were based on the weighted sum of constituent model predictions. Each model prediction was weighted by its F1-score on the validation set in order to assign a higher weight to the model that performed better on the validation set. This weighted sum is then thresholded by the F1-score averaged across models for final prediction. In our study, XLNet and BLOOM-560M received the predominant weights for attaining the highest F1 score on abortion and gun-control datasets respectively.

\subsection{Multimodal Stance Prediction}
To evaluate the utility of image augmentation to text and the possible ways to achieve this, we studied models from different frameworks. The ViLT \cite{kim2021vilt} is a popular vision-language transformer model with reduced computational overhead because of its convolution-free architecture. FLAVA \cite{singh2022flava}, a multimodal model built to generalize to both vision tasks and language tasks. Both models were fine-tuned over the gun control and abortion datasets for the support stance prediction task.

\begin{figure}[h]
    \centering
    \includegraphics[width=\linewidth]{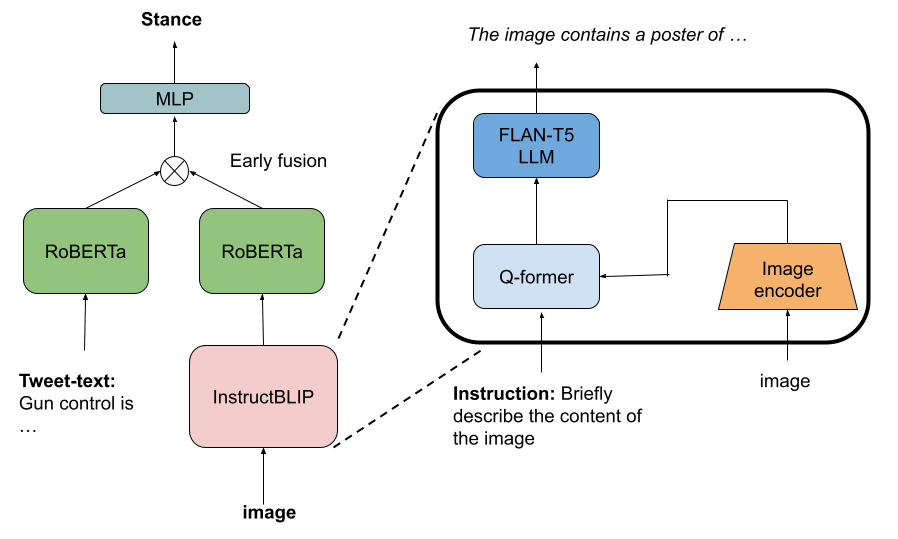}
    \caption{Multimodal RoBERTa configuration. The figure shows the input image summarized as text through instructBLIP and then used to fine-tune the RoBERTa model together with the tweet-text. Shared color between RoBERTa models indicates tied weights.}
    \label{fig:multimodal_roberta}
\end{figure}

Recent vision-language pre-trained models such as instructBLIP \cite{dai2023instructblip} have demonstrated solving image-centric tasks through natural language. We leverage this instruction-based summarization of image content with instructBLIP. Specifically, we summarize each image using the \textit{briefly describe the content of the image} instruction. The resulting textual descriptions of images along with their corresponding tweets were used for stance prediction by fine-tuning a RoBERTa \cite{liu2019roberta} classifier followed by early fusion. We refer to this configuration (Figure \ref{fig:multimodal_roberta}) as the multimodal RoBERTa. 

\subsection{Few-Shot Stance Prediction using LLMs}
Few-shot prediction typically involves using relevant examples during training to learn a new concept that was not included in pretraining. It has been a success not just in conventional language-based tasks but also in multimodal tasks \cite{luo2020univl}. The large and diverse pre-training corpora used in training the foundation models is attributed as one of the reasons for their success in learning with a limited resources paradigm. Using LLaMA-2 \cite{touvron2023llama}, we performed stance prediction in few-shot setting. LLaMA-2 was chosen because of its open-source implementation that outperforms commercial large-scale GPT-3 \cite{brown2020language} with fast inference.

\paragraph{Choice of few-shot examples:} Arguments can be made from different viewpoints or themes. For example, gun control can be referred to from ordinary themes such as the constitutionally granted \textit{right to bear arms, governmental overreach} to targeted themes or experiences such as \textit{school shootings}. We believe that the ImgArg dataset encompasses these diverse themes and wish to leverage the in-context examples that correspond to the same theme for few-shot experiments. We identify the themes in the training set using k-means clustering and pick examples from the same theme cluster during inference. Performance on the validation set was used as a benchmark to identify 12 clusters for the gun control dataset and 13 clusters for the abortion dataset. The manually identified cluster themes are presented as Table \ref{tab:few-shot_clusters} in the Appendix \ref{sec:appendix}. 

\subsection{Experimental Setup}
The imbalance in the abortion dataset is addressed using a weighted cross-entropy loss. Increased weightage was allocated to the minority category loss. The models were trained using HuggingFace \cite{wolf2020huggingfaces} on two A100 NVIDIA GPU environment\footnote{The code is available at: https://github.com/arushi-08/EMNLP-ImageArgTask-PittPixelPersuaders}. Hyperparameters (learning rate, scheduler and weight decay) were optimized for the validation set and performance is reported as precision, recall and F1-score for the support stance and oppose on the test set. More experimental details are shown in Appendix \ref{sec:appendix} section.



\section{Results}

\subsection{Support Stance}

\begin{table}[!h]
  \centering
    \begin{tabular}{cccc}
    \hline
    \textbf{Model} & \textbf{Precision} & \textbf{Recall} & \textbf{F1} \\
    \hline
    XLNet &  0.619 & 0.924& 0.741 \\
    BLOOM-1B & 0.760  & 0.660  & 0.710 \\
    BLOOM-560M &  0.707 & 0.898 & 0.791 \\
    Transformer-XL &   0.571  & 0.881    & 0.693 \\
    DeBERTa-v2 & 0.560      &  0.710     & 0.630 \\
    XLM-RoBERTa &   0.650  & 0.880  & 0.750 \\
    \textbf{Ensemble} & \textbf{0.743} & \textbf{0.906} & \textbf{0.817} \\
    \hline
    \end{tabular}%
\caption{\textit{Support} stance performance using text-based transformer models.}
\label{tab:ensemble-support}%
\end{table}%

Table \ref{tab:ensemble-support} compares the support class performance of individual language models against their ensemble model. The ensemble model used BLOOM-560M as it performed better than its larger counterpart on the validation  set. The constituent models typically have a better recall but low precision, the ensemble model improves precision with a limited drop in the recall. Best performance was observed with the ensemble of unimodal language models with 0.817 F1-score.

\begin{table}[!h]
  \centering
  \resizebox{\columnwidth}{!}{%
    \begin{tabular}{cccc}
    \hline
    \textbf{Model} & \textbf{Precision} & \textbf{Recall} & \textbf{F1} \\
    \hline
    ViLT  & 0.680 & 0.432  & 0.528 \\
    FLAVA &  0.570     &  0.650     & 0.610 \\
    \textbf{Multimodal RoBERTa} & \textbf{0.531} & \textbf{0.932} & \textbf{0.677} \\
    \hline
    \end{tabular}%
  }
  \caption{\textit{Support} stance performance using image-text multimodal transformer models.}
\label{tab:multmodal_results}%
\end{table}%

Multimodal RoBERTa and FLAVA sacrificed precision for recall (shown in Table \ref{tab:multmodal_results}) upon fine-tuning. Both multimodal RoBERTa and FLAVA that leverage images in pixel-space achieve a recall of 0.932 and 0.650 respectively. However, their low precision (0.531 and 0.570 respectively) underperforms the ViLT model. Summarizing images to fine-tune smaller language models tends to result in improved recall albeit at the cost of precision. This approach achieves the highest among the multimodal models with an F1-score of 0.677.

\begin{table}[h]
  \centering
  \resizebox{\columnwidth}{!}{%
    \begin{tabular}{cccc}
    \hline
    \textbf{Model} & \textbf{Precision} & \textbf{Recall} & \textbf{F1} \\
    \hline
    Baseline (\textit{support} only) &   0.395    &  1.000     & 0.566 \\
    zero-shot &   0.440    &  0.290     & 0.350 \\
    four-shot & 0.420  & 0.640  & 0.500 \\
    \textbf{four-shot w/ k-means} & \textbf{0.450} & \textbf{0.700} & \textbf{0.550} \\
    \hline
    \end{tabular}%
  }
\caption{\textit{Support} stance performance using LLaMA-2 based few-shot experiments.}
\label{tab:few-shot_results}%
\end{table}%

We compare our few-shot experiments with the baseline \textit{support} only stance predictions to observe that both zero-shot and four-shot models underperform the baseline. The best performance is demonstrated using the four-shot model with k-means clustering. Clustering was found to improve the recall by 6\% while precision has improved by 3\%. F1-score has improved by 5\% to 0.550. Few-shot LLaMA-2 underperforms the ensemble model at stance prediction.


\subsection{Oppose Stance}
Table \ref{table:ensemble-oppose} shows that the language models have higher precision than recall for the oppose class as compared to the support class (Table \ref{tab:ensemble-support}). Higher precision and lower recall shows us that the text-based language models prioritize predicting the support stance (minority class).
Moreover, the ensemble approach outperforms other language models even on the oppose stance. 
For the multimodal models, both the ViLT and FLAVA models demonstrated superior performance for the oppose class (shown in Table \ref{tab:multmodal_results_oppose}) compared to the support class (shown in Table \ref{tab:multmodal_results}). However, the multimodal RoBERTa model follows similar pattern as text-based language models, in terms of scoring high on recall for support class vs oppose class. For LLaMa-2 experiments, The F1 scores for the support class (Table \ref{tab:few-shot_results}) across all methods are consistently higher compared to the oppose class (Table \ref{tab:few-shot_results_oppose}). This suggests that LLaMa-2 is more adept at discerning patterns associated with the support class than those of the oppose class.

\begin{table}[H]
  \centering
  \resizebox{\columnwidth}{!}{%
    \begin{tabular}{cccc}
    \hline
    \textbf{Model} & \textbf{Precision} & \textbf{Recall} & \textbf{F1} \\
    \hline
    XLNet & 0.927 & 0.630 & 0.750 \\
    Bloom-1B & 0.790  & 0.870  & 0.830 \\
    Bloom-560M & 0.919  &   0.757   &  0.770\\
    Transformer-XL   &  0.880  & 0.569 &   0.691 \\
    DeBERTa-v2 & 0.770      &  0.640     & 0.700 \\
    XLM-RoBERTa &  0.691    &  0.899   & 0.781  \\
    \textbf{Ensemble} & \textbf{0.929} & \textbf{0.796} & \textbf{0.857} \\
    \hline
    \end{tabular}%
}    
\caption{\textit{Oppose} stance performance using text-based transformer models.}
\label{table:ensemble-oppose}%
\end{table}%

\begin{table}[]
  \centering
  \resizebox{\columnwidth}{!}{%
    \begin{tabular}{cccc}
    \hline
    \textbf{Model} & \textbf{Precision} & \textbf{Recall} & \textbf{F1} \\
    \hline
    ViLT  & 0.701  &   0.867  &   0.775 \\
    FLAVA &  0.750     &    0.690   &  0.720\\
    Multimodal RoBERTa & 0.913 & 0.464 & 0.615 \\
    \hline
    \end{tabular}%
  }
  \caption{\textit{Oppose} stance performance using image-text multimodal transformer models.}
  \label{tab:multmodal_results_oppose}%
\end{table}%


\begin{table}[H]
  \centering
  \resizebox{\columnwidth}{!}{%
    \begin{tabular}{cccc}
    \hline
    \textbf{Model} & \textbf{Precision} & \textbf{Recall} & \textbf{F1} \\
    \hline
    Baseline (\textit{Oppose} only) &   0.605    &  1.000    &  0.754 \\    
    zero-shot & 0.690   &  0.060 & 0.110 \\ 
    four-shot & 0.770 & 0.300  & 0.430 \\   
    four-shot w/ k-means & 0.740 & 0.270 & 0.400 \\
    \hline
    \end{tabular}%
  
  }
\caption{\textit{Oppose} stance performance using LLaMA-2 based few-shot experiments.}
\label{tab:few-shot_results_oppose}%
\end{table}%

\section{Discussion}

Popular pre-trained language models such as XLNet, BLOOM, Transformer-XL, DeBERTa-v2 and XLM-RoBERTa were fine-tuned for stance prediction on tweets about gun control and abortion. Results demonstrate that the ensemble of these models performs better than any of the constituent models. However, the disparity is limited. XLNet achieves better recall than the ensemble model and similarly, the BLOOM-560M underperforms the ensemble by 0.026 (though precision of the ensemble is higher by 0.036).  This raises the trade-off question between ensemble performance vs. the large computational requirement justified for marginal improvement in the performance.

The best performing multimodal model used the image content summarized as text, unlike its counterpart models that operate in pixel space. We believe the diversity of the images contributes to this difference. In addition to typical images containing people and objects such as guns, trucks and so on, the training set also contained propaganda-related material such as posters with statements. While vision-language models are increasingly getting better at object-centric tasks, understanding such material is closely related to problems such as optical character recognition, which are not often explored in pretraining vision-language models. Our instruction-based image summarization suggests that when explicitly prompted, vision-language models excel not just at object-centric descriptions of images but also at recognizing text from images. Attempts were made to incorporate demographic factors such as number of people in the image, their skin color and gender. However, manual inspection revealed that the resultant instructBLIP predictions were not reliable. Despite augmenting language modality with images in different ways, text-based models outperformed the multimodal models. 



Out-of-the-box LLaMA-2 underperforms the baseline \textit{support} only prediction model. However, prompting through four-shot examples greatly improves the performance. This is further enhanced by using in-context examples. This demonstrates that in-context examples that potentially share similar theme (not necessarily the stance) tend to capture the stance better than arbitrary examples from the dataset. The themes were found to include discussions along mental health, effects on children, racism, illegal acquisition, etc. for the gun control dataset; Supreme Court, birth control, religion, reproductive rights, etc. for the abortion dataset.

\section{Conclusions and Future Work}

Our investigation questions the necessity of images to predict stance in multimodal tweets through different ways of using image-based information in conjunction with text-based language models and investigating the inherent capabilities of LLMs for stance prediction. Results suggest that the best performance can be achieved using an ensemble of language models. Our experiments with multimodal models do not completely refute the utility of images for stance prediction, rather they merely evaluate the current state-of-the-art multimodal models. Incorporating domain knowledge \cite{lewis2021retrievalaugmented}, and alternative prompting methods like Question Decomposition \cite{radhakrishnan2023question} and Tree-of-Thought \cite{yao2023tree} which provide the rationale for the prediction in addition to the stance provide a future direction to address the limited performance with LLaMA-2.

\bibliography{emnlp2023}

\begin{thebibliography}{30}
\expandafter\ifx\csname natexlab\endcsname\relax\def\natexlab#1{#1}\fi

\bibitem[{Alturayeif et~al.(2023)Alturayeif, Luqman, and
  Ahmed}]{alturayeif2023systematic}
Nora Alturayeif, Hamzah Luqman, and Moataz Ahmed. 2023.
\newblock A systematic review of machine learning techniques for stance
  detection and its applications.
\newblock \emph{Neural Computing and Applications}, 35(7):5113--5144.

\bibitem[{Brown et~al.(2020)Brown, Mann, Ryder, Subbiah, Kaplan, Dhariwal,
  Neelakantan, Shyam, Sastry, Askell, Agarwal, Herbert-Voss, Krueger, Henighan,
  Child, Ramesh, Ziegler, Wu, Winter, Hesse, Chen, Sigler, Litwin, Gray, Chess,
  Clark, Berner, McCandlish, Radford, Sutskever, and
  Amodei}]{brown2020language}
Tom~B. Brown, Benjamin Mann, Nick Ryder, Melanie Subbiah, Jared Kaplan,
  Prafulla Dhariwal, Arvind Neelakantan, Pranav Shyam, Girish Sastry, Amanda
  Askell, Sandhini Agarwal, Ariel Herbert-Voss, Gretchen Krueger, Tom Henighan,
  Rewon Child, Aditya Ramesh, Daniel~M. Ziegler, Jeffrey Wu, Clemens Winter,
  Christopher Hesse, Mark Chen, Eric Sigler, Mateusz Litwin, Scott Gray,
  Benjamin Chess, Jack Clark, Christopher Berner, Sam McCandlish, Alec Radford,
  Ilya Sutskever, and Dario Amodei. 2020.
\newblock \href {http://arxiv.org/abs/2005.14165} {Language models are few-shot
  learners}.

\bibitem[{Conneau et~al.(2019)Conneau, Khandelwal, Goyal, Chaudhary, Wenzek,
  Guzm{\'a}n, Grave, Ott, Zettlemoyer, and Stoyanov}]{conneau2019xlm-roberta}
Alexis Conneau, Kartikay Khandelwal, Naman Goyal, Vishrav Chaudhary, Guillaume
  Wenzek, Francisco Guzm{\'a}n, Edouard Grave, Myle Ott, Luke Zettlemoyer, and
  Veselin Stoyanov. 2019.
\newblock Unsupervised cross-lingual representation learning at scale.
\newblock \emph{arXiv preprint arXiv:1911.02116}.

\bibitem[{Dai et~al.(2023)Dai, Li, Li, Tiong, Zhao, Wang, Li, Fung, and
  Hoi}]{dai2023instructblip}
Wenliang Dai, Junnan Li, Dongxu Li, Anthony Meng~Huat Tiong, Junqi Zhao,
  Weisheng Wang, Boyang Li, Pascale Fung, and Steven Hoi. 2023.
\newblock \href {http://arxiv.org/abs/2305.06500} {Instructblip: Towards
  general-purpose vision-language models with instruction tuning}.

\bibitem[{Dai et~al.(2019)Dai, Yang, Yang, Carbonell, Le, and
  Salakhutdinov}]{dai2019transformer-xl}
Zihang Dai, Zhilin Yang, Yiming Yang, Jaime Carbonell, Quoc~V Le, and Ruslan
  Salakhutdinov. 2019.
\newblock Transformer-xl: Attentive language models beyond a fixed-length
  context.
\newblock \emph{arXiv preprint arXiv:1901.02860}.

\bibitem[{Gugger et~al.(2022)Gugger, Debut, Wolf, Schmid, Mueller, Mangrulkar,
  Sun, and Bossan}]{accelerate}
Sylvain Gugger, Lysandre Debut, Thomas Wolf, Philipp Schmid, Zachary Mueller,
  Sourab Mangrulkar, Marc Sun, and Benjamin Bossan. 2022.
\newblock Accelerate: Training and inference at scale made simple, efficient
  and adaptable.
\newblock \url{https://github.com/huggingface/accelerate}.

\bibitem[{He et~al.(2020)He, Liu, Gao, and Chen}]{he2020deberta}
Pengcheng He, Xiaodong Liu, Jianfeng Gao, and Weizhu Chen. 2020.
\newblock Deberta: Decoding-enhanced bert with disentangled attention.
\newblock \emph{arXiv preprint arXiv:2006.03654}.

\bibitem[{Hosseinia et~al.(2020)Hosseinia, Dragut, and
  Mukherjee}]{hosseinia2020stance}
Marjan Hosseinia, Eduard Dragut, and Arjun Mukherjee. 2020.
\newblock Stance prediction for contemporary issues: Data and experiments.
\newblock \emph{arXiv preprint arXiv:2006.00052}.

\bibitem[{Jiang et~al.(2023)Jiang, Ren, and Lin}]{jiang2023llm}
Dongfu Jiang, Xiang Ren, and Bill~Yuchen Lin. 2023.
\newblock Llm-blender: Ensembling large language models with pairwise ranking
  and generative fusion.
\newblock \emph{arXiv preprint arXiv:2306.02561}.

\bibitem[{Kim et~al.(2021)Kim, Son, and Kim}]{kim2021vilt}
Wonjae Kim, Bokyung Son, and Ildoo Kim. 2021.
\newblock Vilt: Vision-and-language transformer without convolution or region
  supervision.
\newblock In \emph{International Conference on Machine Learning}, pages
  5583--5594. PMLR.

\bibitem[{Lewis et~al.(2021)Lewis, Perez, Piktus, Petroni, Karpukhin, Goyal,
  Küttler, Lewis, tau Yih, Rocktäschel, Riedel, and
  Kiela}]{lewis2021retrievalaugmented}
Patrick Lewis, Ethan Perez, Aleksandra Piktus, Fabio Petroni, Vladimir
  Karpukhin, Naman Goyal, Heinrich Küttler, Mike Lewis, Wen tau Yih, Tim
  Rocktäschel, Sebastian Riedel, and Douwe Kiela. 2021.
\newblock \href {http://arxiv.org/abs/2005.11401} {Retrieval-augmented
  generation for knowledge-intensive nlp tasks}.

\bibitem[{Liu et~al.(2019)Liu, Ott, Goyal, Du, Joshi, Chen, Levy, Lewis,
  Zettlemoyer, and Stoyanov}]{liu2019roberta}
Yinhan Liu, Myle Ott, Naman Goyal, Jingfei Du, Mandar Joshi, Danqi Chen, Omer
  Levy, Mike Lewis, Luke Zettlemoyer, and Veselin Stoyanov. 2019.
\newblock \href {http://arxiv.org/abs/1907.11692} {Roberta: A robustly
  optimized bert pretraining approach}.

\bibitem[{Liu et~al.(2023)Liu, Elaraby, Zhong, and
  Litman}]{liu-etal-2023-overview}
Zhexiong Liu, Mohamed Elaraby, Yang Zhong, and Diane Litman. 2023.
\newblock Overview of {I}mage{A}rg-2023: The first shared task in multimodal
  argument mining.
\newblock In \emph{Proceedings of the 10th Workshop on Argument Mining}, Online
  and in Singapore. Association for Computational Linguistics.

\bibitem[{Liu et~al.(2022)Liu, Guo, Dai, and Litman}]{liu2022imagearg}
Zhexiong Liu, Meiqi Guo, Yue Dai, and Diane Litman. 2022.
\newblock Imagearg: A multi-modal tweet dataset for image persuasiveness
  mining.
\newblock \emph{arXiv preprint arXiv:2209.06416}.

\bibitem[{Loshchilov and Hutter(2017)}]{DBLP:journals/corr/abs-1711-05101}
Ilya Loshchilov and Frank Hutter. 2017.
\newblock \href {http://arxiv.org/abs/1711.05101} {Fixing weight decay
  regularization in adam}.
\newblock \emph{CoRR}, abs/1711.05101.

\bibitem[{Luo et~al.(2020)Luo, Ji, Shi, Huang, Duan, Li, Li, Bharti, and
  Zhou}]{luo2020univl}
Huaishao Luo, Lei Ji, Botian Shi, Haoyang Huang, Nan Duan, Tianrui Li, Jason
  Li, Taroon Bharti, and Ming Zhou. 2020.
\newblock Univl: A unified video and language pre-training model for multimodal
  understanding and generation.
\newblock \emph{arXiv preprint arXiv:2002.06353}.

\bibitem[{Radhakrishnan et~al.(2023)Radhakrishnan, Nguyen, Chen, Chen, Denison,
  Hernandez, Durmus, Hubinger, Kernion, Lukošiūtė, Cheng, Joseph, Schiefer,
  Rausch, McCandlish, Showk, Lanham, Maxwell, Chandrasekaran, Hatfield-Dodds,
  Kaplan, Brauner, Bowman, and Perez}]{radhakrishnan2023question}
Ansh Radhakrishnan, Karina Nguyen, Anna Chen, Carol Chen, Carson Denison, Danny
  Hernandez, Esin Durmus, Evan Hubinger, Jackson Kernion, Kamilė Lukošiūtė,
  Newton Cheng, Nicholas Joseph, Nicholas Schiefer, Oliver Rausch, Sam
  McCandlish, Sheer~El Showk, Tamera Lanham, Tim Maxwell, Venkatesa
  Chandrasekaran, Zac Hatfield-Dodds, Jared Kaplan, Jan Brauner, Samuel~R.
  Bowman, and Ethan Perez. 2023.
\newblock \href {http://arxiv.org/abs/2307.11768} {Question decomposition
  improves the faithfulness of model-generated reasoning}.

\bibitem[{Reveilhac and Schneider(2023)}]{reveilhac2023replicable}
Maud Reveilhac and Gerold Schneider. 2023.
\newblock Replicable semi-supervised approaches to state-of-the-art stance
  detection of tweets.
\newblock \emph{Information Processing \& Management}, 60(2):103199.

\bibitem[{Scao et~al.(2022)Scao, Fan, Akiki, Pavlick, Ili{\'c}, Hesslow,
  Castagn{\'e}, Luccioni, Yvon, Gall{\'e} et~al.}]{scao2022bloom}
Teven~Le Scao, Angela Fan, Christopher Akiki, Ellie Pavlick, Suzana Ili{\'c},
  Daniel Hesslow, Roman Castagn{\'e}, Alexandra~Sasha Luccioni, Fran{\c{c}}ois
  Yvon, Matthias Gall{\'e}, et~al. 2022.
\newblock Bloom: A 176b-parameter open-access multilingual language model.
\newblock \emph{arXiv preprint arXiv:2211.05100}.

\bibitem[{Singh et~al.(2022)Singh, Hu, Goswami, Couairon, Galuba, Rohrbach, and
  Kiela}]{singh2022flava}
Amanpreet Singh, Ronghang Hu, Vedanuj Goswami, Guillaume Couairon, Wojciech
  Galuba, Marcus Rohrbach, and Douwe Kiela. 2022.
\newblock Flava: A foundational language and vision alignment model.
\newblock In \emph{Proceedings of the IEEE/CVF Conference on Computer Vision
  and Pattern Recognition}, pages 15638--15650.

\bibitem[{Sobhani(2017)}]{sobhani2017stance}
Parinaz Sobhani. 2017.
\newblock \emph{Stance detection and analysis in social media}.
\newblock Ph.D. thesis, Universite d'Ottawa/University of Ottawa.

\bibitem[{Thorndike(1953)}]{Thorndike1953WhoBI}
Robert~L. Thorndike. 1953.
\newblock \href {https://api.semanticscholar.org/CorpusID:120467216} {Who
  belongs in the family?}
\newblock \emph{Psychometrika}, 18:267--276.

\bibitem[{Touvron et~al.(2023)Touvron, Lavril, Izacard, Martinet, Lachaux,
  Lacroix, Rozière, Goyal, Hambro, Azhar, Rodriguez, Joulin, Grave, and
  Lample}]{touvron2023llama}
Hugo Touvron, Thibaut Lavril, Gautier Izacard, Xavier Martinet, Marie-Anne
  Lachaux, Timothée Lacroix, Baptiste Rozière, Naman Goyal, Eric Hambro,
  Faisal Azhar, Aurelien Rodriguez, Armand Joulin, Edouard Grave, and Guillaume
  Lample. 2023.
\newblock \href {http://arxiv.org/abs/2302.13971} {Llama: Open and efficient
  foundation language models}.

\bibitem[{Upadhyaya et~al.(2023)Upadhyaya, Fisichella, and
  Nejdl}]{Upadhyaya2023AMM}
Apoorva Upadhyaya, Marco Fisichella, and Wolfgang Nejdl. 2023.
\newblock \href {https://api.semanticscholar.org/CorpusID:258333733} {A
  multi-task model for emotion and offensive aided stance detection of climate
  change tweets}.
\newblock \emph{Proceedings of the ACM Web Conference 2023}.

\bibitem[{Wang et~al.(2023)Wang, Chen, Qian, Gao, Wei, Wang, Tian, and
  Gao}]{Wang2023LargescaleMP}
Xiao Wang, Guangyao Chen, Guangwu Qian, Pengcheng Gao, Xiaoyong Wei, Yaowei
  Wang, Yonghong Tian, and Wen Gao. 2023.
\newblock \href {https://api.semanticscholar.org/CorpusID:257038341}
  {Large-scale multi-modal pre-trained models: A comprehensive survey}.
\newblock \emph{Machine Intelligence Research}, 20:447 -- 482.

\bibitem[{Wolf et~al.(2020)Wolf, Debut, Sanh, Chaumond, Delangue, Moi, Cistac,
  Rault, Louf, Funtowicz, Davison, Shleifer, von Platen, Ma, Jernite, Plu, Xu,
  Scao, Gugger, Drame, Lhoest, and Rush}]{wolf2020huggingfaces}
Thomas Wolf, Lysandre Debut, Victor Sanh, Julien Chaumond, Clement Delangue,
  Anthony Moi, Pierric Cistac, Tim Rault, Rémi Louf, Morgan Funtowicz, Joe
  Davison, Sam Shleifer, Patrick von Platen, Clara Ma, Yacine Jernite, Julien
  Plu, Canwen Xu, Teven~Le Scao, Sylvain Gugger, Mariama Drame, Quentin Lhoest,
  and Alexander~M. Rush. 2020.
\newblock \href {http://arxiv.org/abs/1910.03771} {Huggingface's transformers:
  State-of-the-art natural language processing}.

\bibitem[{Yang et~al.(2019)Yang, Dai, Yang, Carbonell, Salakhutdinov, and
  Le}]{yang2019xlnet}
Zhilin Yang, Zihang Dai, Yiming Yang, Jaime Carbonell, Russ~R Salakhutdinov,
  and Quoc~V Le. 2019.
\newblock Xlnet: Generalized autoregressive pretraining for language
  understanding.
\newblock \emph{Advances in neural information processing systems}, 32.

\bibitem[{Yao et~al.(2023)Yao, Yu, Zhao, Shafran, Griffiths, Cao, and
  Narasimhan}]{yao2023tree}
Shunyu Yao, Dian Yu, Jeffrey Zhao, Izhak Shafran, Thomas~L. Griffiths, Yuan
  Cao, and Karthik Narasimhan. 2023.
\newblock \href {http://arxiv.org/abs/2305.10601} {Tree of thoughts: Deliberate
  problem solving with large language models}.

\bibitem[{Zadeh et~al.(2018)Zadeh, Liang, Morency, Poria, Cambria, and
  Scherer}]{zadeh2018proceedings}
Amir Zadeh, Paul~Pu Liang, Louis-Philippe Morency, Soujanya Poria, Erik
  Cambria, and Stefan Scherer. 2018.
\newblock Proceedings of grand challenge and workshop on human multimodal
  language (challenge-hml).
\newblock In \emph{Proceedings of Grand Challenge and Workshop on Human
  Multimodal Language (Challenge-HML)}.

\bibitem[{Zhang et~al.(2021)Zhang, Qian, Fang, and Xu}]{Zhang2021MultiModalMM}
Huaiwen Zhang, Shengsheng Qian, Quan Fang, and Changsheng Xu. 2021.
\newblock \href {https://api.semanticscholar.org/CorpusID:234272644}
  {Multi-modal meta multi-task learning for social media rumor detection}.
\newblock \emph{IEEE Transactions on Multimedia}, 24:1449--1459.

\end{thebibliography}
\bibliographystyle{acl_natbib}

\appendix

\section{Appendix}
\label{sec:appendix}
This appendix provides details such as the number of parameters in the final classification, hyperparameters and finetuning approach for various models\footnote{code used in this work would be made available after the review process to preserve the anonymity of the authors} in this work. All models used the Adam \cite{DBLP:journals/corr/abs-1711-05101} optimizer.

\begin{table}[!h]
\centering
\resizebox{\columnwidth}{!}{%
\begin{tabular}{|l|l|}
\hline
\textbf{Model}          & \textbf{Size of classification head} \\ \hline
XLNet          & 1024 \\ \hline
Bloom-1B       & 64   \\ \hline
Bloom-560M     & 64   \\ \hline
Transformer-XL & 1024 \\ \hline
DeBERTa-v2     & 1536 \\ \hline
XLM-RoBERTa    & 768  \\ \hline
Multimodal RoBERTa (MLP)    &  1536  \\ \hline
FLAVA   &  768  \\ \hline
ViLT  &  768  \\ \hline
\end{tabular}
}
\caption{Table showing the size of the final classification layer for various models used in this work.}
\label{tab:final_layer}
\end{table}

\subsection{Ensemble Stance Prediction Model}

We employed various pretrained language models, specifically {XLNet}\footnote{\url{https://huggingface.co/xlnet-base-cased}}, {BLOOM-560M}\footnote{\url{https://huggingface.co/bigscience/bloom-560m}}, {Transformer-XL}\footnote{\url{https://huggingface.co/transfo-xl-wt103}}, {DeBERTa-v2}\footnote{\url{https://huggingface.co/microsoft/deberta-v2-xlarge}}, and {XLM-RoBeRTa}\footnote{\url{https://huggingface.co/facebook/xlm-roberta-xl}}. Each model was augmented with a classification head for binary sequence classification tasks. The summary of the size of the classification head for each model is provided in Table \ref{tab:final_layer}. We utilized Adam optimizer with a learning rate of 1e-3. A learning rate scheduler was also incorporated into the training regimen with a patience of 3. To mitigate the risk of model overfitting, a weight decay parameter was set at 0.01. All models were trained for 10 epochs. 


\subsection{Multimodal Stance Prediction Model}

\begin{table*}[t]
  \centering
    \begin{tabular}{cccc}
\cline{1-2}    \textbf{Gun control} & \textbf{Abortion} &       &  \\
\cline{1-2}    Gun violence as a mental health problem & Natural Law Right to Life &       &  \\
    Effects of gun violence on children & Abortion is evil &       &  \\
    Pro-gun control politicians & Supreme Court and abortion &       &  \\
    Racism and gun control & Abortion is murder &       &  \\
    Trump and guns & Birth control pills &       &  \\
    Illegal acquisition of guns & Pro-life &       &  \\
    Supreme Court and gun control & Religion and motherhood &       &  \\
    Second amendment right & Reproductive rights of women &       &  \\
     & \#savethebabyhumans hashtag &       &  \\
          & Roe v. Wade abortion case &       &  \\
\cline{1-2}    \end{tabular}%
  \caption{Themes identified using k-means clustering for few-shot examples in gun control and abortion datasets.Same theme(s) captured by multiple clusters resulted in fewer themes than reported clusters.}
  \label{tab:few-shot_clusters}%
\end{table*}%

For the multimodal {RoBERTa} \footnote{\url{https://huggingface.co/roberta-base}}, the learning rate was configured at 5e-2, and the weight decay parameter was set at 0.01 during the fine-tuning process. The training continued until the validation loss ceased to decrease for five consecutive epochs. Figure \ref{fig:multimodal_roberta} presented the visualization of the Multimodal RoBERTa. For the {ViLT}\footnote{\url{https://huggingface.co/docs/transformers/model_doc/vilt}} model, a low learning of 2.25e-6 was found to be optimal. The model underwent training for a total of 10 epochs.
In the case of the {FLAVA}\footnote{\url{https://huggingface.co/facebook/flava-full}} model, an early stopping mechanism was implemented, resulting model was trained for six epochs prior to any increase in validation loss. The learning rate for this model was set at 5e-5.

\subsection{Few-shot Stance Prediction Model}

\begin{figure}[H]
    \centering
    \includegraphics[width=0.5\textwidth]{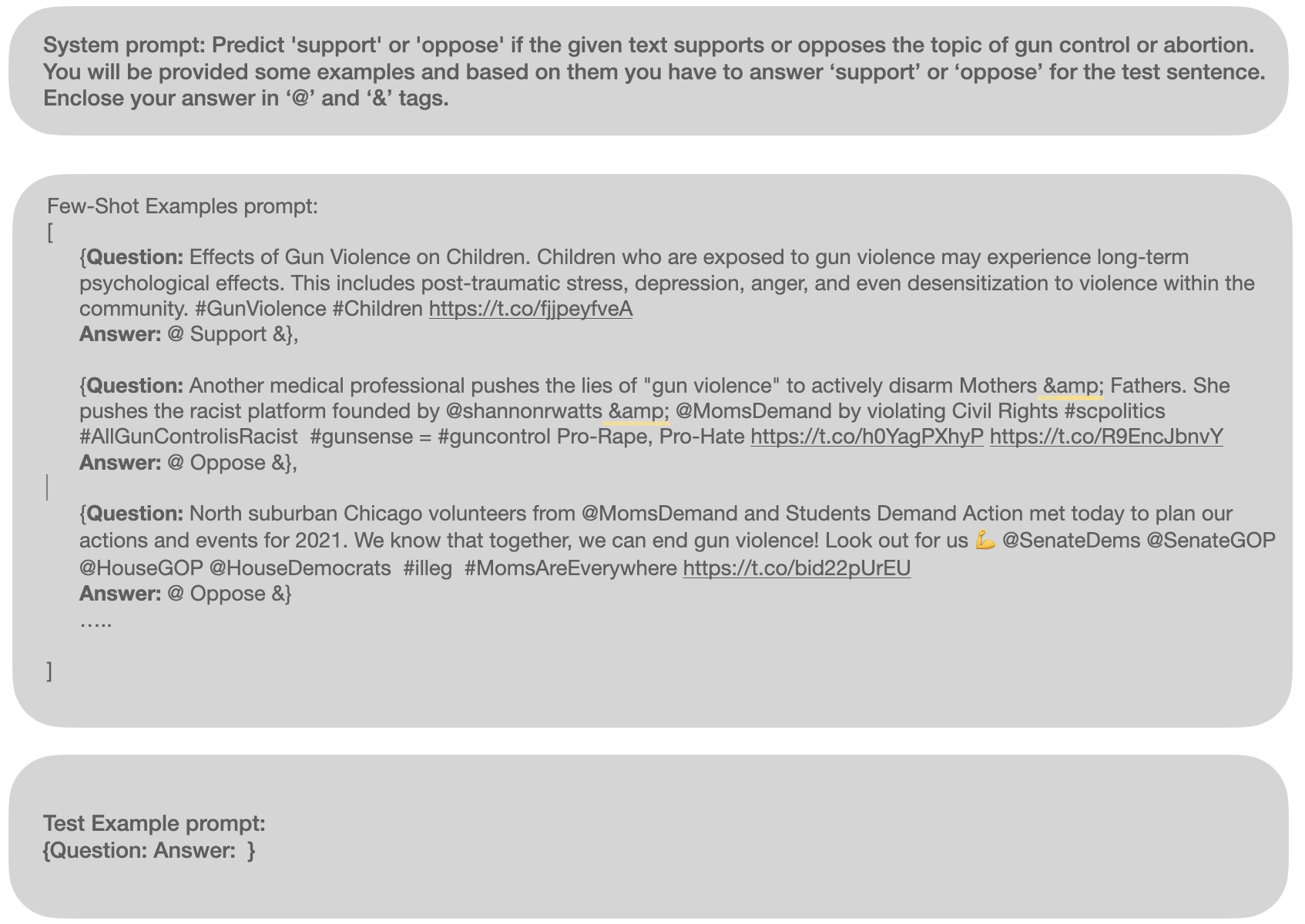}
    \caption{The provided illustration depicts a k-means few-shot prompt template employed in our experimental investigations conducted on the gun control dataset. A comparable configuration was also applied when examining the abortion dataset. For conciseness, we have omitted the inclusion of all four examples in this presentation.}
    \label{fig:few_shot_screenshot}
\end{figure}

In this study, we employed the Hugging Face's {LLaMa-2 13B}\footnote{\url{https://huggingface.co/meta-llama/Llama-2-13b-hf}} model for inference, leveraging the capabilities of Hugging Face Accelerate \cite{accelerate}. 
The experimental design utilized {Langchain}\footnote{\url{https://github.com/langchain-ai/langchain}}  to formulate a tripartite template for prompt engineering. The template is segmented into three distinct components: The system prompt, which serves as a generic instructional scaffold for the language model, a set of few-shot examples to guide the model's responses, and the test set tweet that the model is tasked to analyze. While the standard convention of using no examples for zero-shot and sampling four arbitrary examples for four-shot prediction was used, in the four-shot with k-means, the training set is initially partitioned into clusters using the k-means algorithm (12 clusters for gun control and 13 for abortion). For each test example, its corresponding cluster is predicted, and four examples are randomly sampled from the cluster as few-shot examples. The optimal number of clusters was ascertained using the Elbow Method \cite{Thorndike1953WhoBI}. Table \ref{tab:few-shot_clusters} presents some prominent themes found using k-means clustering in gun control and abortion datasets. For LLM output generation, the temperature parameter was set to zero, and the 'top\_k' parameter was configured at 30. We employed a Multinomial sampling strategy, setting the do\_sample = True and num\_beams parameter to 1. An exemplar of the prompt template employed is depicted in Figure \ref{fig:few_shot_screenshot}.

\end{document}